%% file: main.tex
\documentclass[10pt,twocolumn,letterpaper]{article}

\usepackage[pagenumbers]{iccv} 


\definecolor{iccvblue}{rgb}{0.21,0.49,0.74}
\usepackage[pagebackref,breaklinks,colorlinks,allcolors=iccvblue]{hyperref}

\usepackage{colortbl}
\usepackage{adjustbox}

\usepackage[capitalize]{cleveref}
\definecolor{Gray}{gray}{0.93}
\definecolor{Red}{RGB}{255, 46, 23}
\definecolor{Green}{RGB}{0, 171, 79}
\definecolor{cred}{rgb}{0.85, 0.1, 0.15}
\definecolor{cgreen}{rgb}{0.25, 0.68, 0.28}
\definecolor{royalblue}{rgb}{0.25, 0.5, 0.75}
\definecolor{grayblue}{rgb}{0.9, 0.92, 0.95}
\hypersetup{
colorlinks=true,
linkcolor=red,
urlcolor=magenta
}
\usepackage{ulem}
\usepackage{tabulary}
\newcolumntype{I}{!{\vrule width 1pt}}

\newcolumntype{x}[1]{>{\centering\arraybackslash}p{#1pt}}
\newcolumntype{y}[1]{>{\raggedright\arraybackslash}p{#1pt}}
\newcolumntype{z}[1]{>{\raggedleft\arraybackslash}p{#1pt}}

\newcommand\blfootnote[1]{%
  \begingroup
  \renewcommand\thefootnote{}\footnote{#1}%
  \addtocounter{footnote}{-1}%
  \endgroup
}

\title{Multimodal Long Video Modeling Based on Temporal Dynamic Context}

\author{Haoran Hao$^{1,2*}$, Jiaming Han$^{1*}$, Yiyuan Zhang$^{1}$, Xiangyu Yue$^{1\dag}$ \vspace{0.2cm}\\
$^{1}$MMLab, The Chinese University of Hong Kong\quad  $^{2}$Nanjing University
}

\begin{document}
\maketitle
\blfootnote{$^*$ Equal contribution\ \ $^{\dagger}$ Corresponding author}

\input{sec/0_abstract}    
\input{sec/1_intro}
\input{sec/X_suppl}

{
    \small
    \bibliographystyle{ieeenat_fullname}
    \bibliography{main}
}


\end{document}

%% file: sec/0_abstract.tex
\begin{abstract}
Recent advances in Large Language Models (LLMs) have led to significant breakthroughs in video understanding. However, existing models still struggle with long video processing due to the context length constraint of LLMs and the vast amount of information within the video. Although some recent methods are designed for long video understanding, they often lose crucial information during token compression and struggle with additional modality like audio.
\textbf{In this work}, we propose a dynamic long video encoding method utilizing the temporal relationship between frames, named \textbf{Temporal Dynamic Context} (TDC).
\textbf{Firstly}, we segment the video into semantically consistent scenes based on inter-frame similarities, then encode each frame into tokens using visual-audio encoders. \textbf{Secondly}, we propose a novel temporal context compressor to reduce the number of tokens within each segment. Specifically, we employ a query-based Transformer to aggregate video, audio, and instruction text tokens into a limited set of temporal context tokens. \textbf{Finally}, we feed the static frame tokens and the temporal context tokens into the LLM for video understanding.
\textbf{Furthermore}, to handle extremely long videos, we propose a training-free chain-of-thought strategy that progressively extracts answers from multiple video segments. These intermediate answers serve as part of the reasoning process and contribute to the final answer.
We conduct extensive experiments on general video understanding and audio-video understanding benchmarks, where our method demonstrates strong performance.
The code and models are available at \url{https://github.com/Hoar012/TDC-Video}.
\end{abstract}

%% file: sec/1_intro.tex



\begin{figure}[t]
  \centering
  \includegraphics[width=\linewidth]{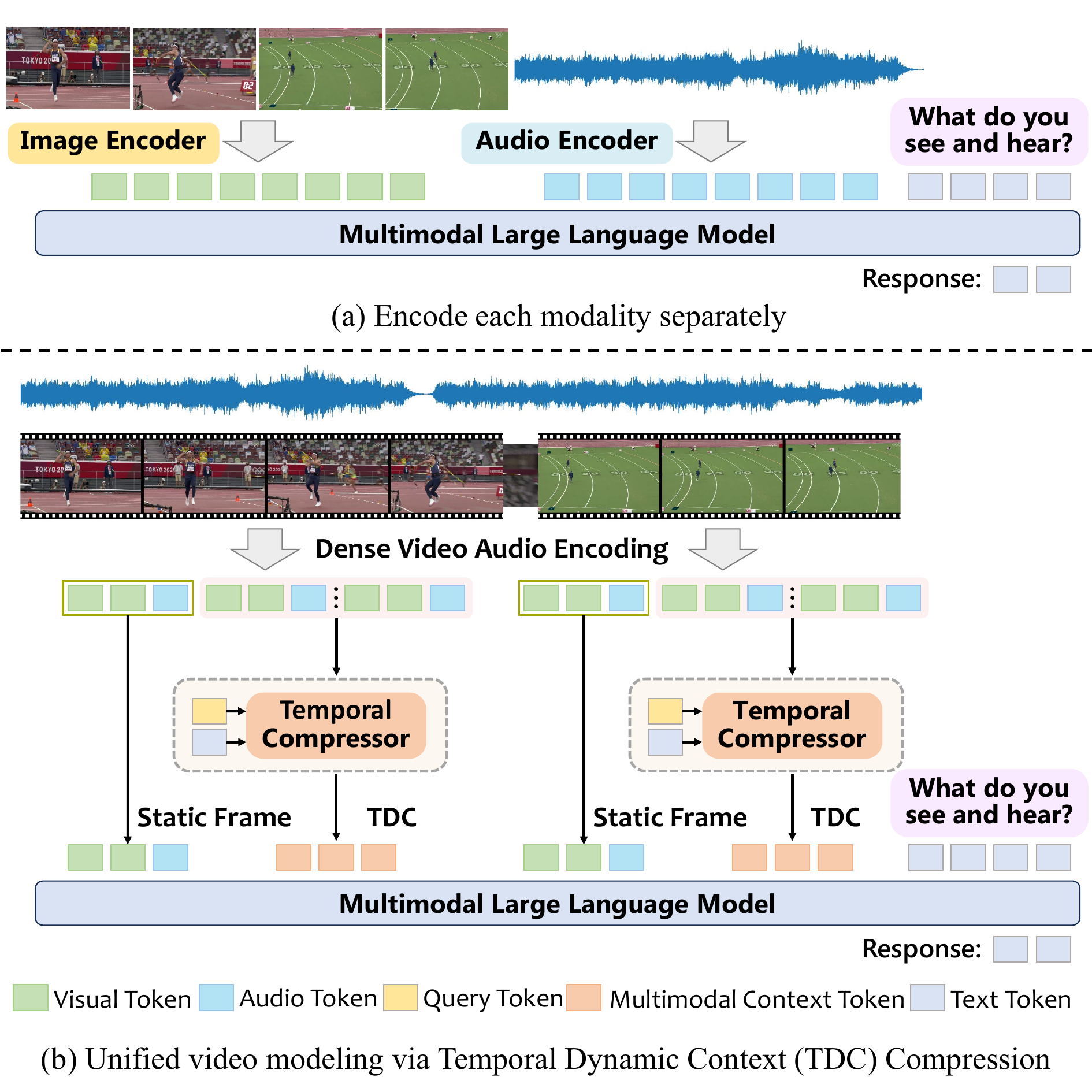}
  \vspace{-0.1cm}
  \caption{\textbf{Comparison of Visual and Audio Encoding in Video Modeling.} (a) Existing methods encode each modality separately and then concatenate them, leading to inconsistencies and difficulties in handling long videos. (b) We propose Temporal Dynamic Context (TDC) compression, which incorporates both static visual features and dynamic video context to represent videos more effectively. This approach enables better multimodal integration and efficient compression for long videos.}
  \label{fig:intro1collum}
\end{figure}

\section{Introduction}
\label{sec:intro}

Recently, advances in large language models (LLMs) \cite{gpt4, llama3-2, qwen2} have significantly improved their ability in language processing and generation. Researchers have extended these models to other modalities, such as vision \cite{llava, llama-adapter, li2023videochat, llavaonevision}, audio \cite{cheng2024videollama2, geng2024longvale} and point clouds \cite{onellm, han2023imagebind-llm}, leading to the development of powerful multimodal LLMs (MLLMs). These MLLMs achieve strong performance in various tasks, such as image captioning \cite{chen2024sharegpt4v, hao2024rap} and question answering \cite{llava, tong2024cambrian1}. However, video understanding remains a challenging problem due to the interplay of multiple modalities \cite{cheng2024videollama2, geng2024longvale} and the complexity of large-scale information \cite{moviechat, li2024llama-vid}, especially in long videos.

A key challenge in long video processing is efficiently representing videos to minimize redundancy between frames while preserving crucial details. Some attempts have been made to address this challenge~\cite{longvlm, longvu, li2024videochat-flash, wang2025internvideo2-5}, but they typically compress video tokens based on visual similarity between frames, while overlooking high-level dynamic semantic information. This limits the model's ability to capture comprehensive video information within a fixed number of tokens, increasing the difficulty of understanding the content.

Another challenge lies in integrating multiple modalities for comprehensive video understanding. For instance, when watching a movie, humans naturally integrate speech, background music, visual scenes, and subtitles to grasp the full context. However, most existing video LLMs are limited to visual and textual modalities and struggle to tackle audio. Although some works~\cite{zhang2023videollama, cheng2024videollama2, geng2024longvale} represented by Video-LLaMA \cite{zhang2023videollama} try to incorporate both audio and visual information, their approach of simply concatenating modality tokens often leads to suboptimal performance by treating different modalities separately (refer to Fig.~\ref{fig:intro1collum} (a)). Developing a unified representation method that effectively connects information between modalities is essential to improve multimodal video understanding.

To address these challenges, we propose a video representation model that integrates multiple modalities within a unified video context and achieves effective token compression. As shown in Fig.~\ref{fig:intro1collum} (b), (1) we segment the video into scenes based on the visual consistency between frames, which are then encoded separately. (2) Each video clip is represented using both static visual features of the key frame and dynamic video context of the video. We first extract per-second features using a visual encoder and an audio encoder. The feature of the first frame is fully retained as a static representation, while the features of subsequent frames are compressed by a Q-Former \cite{instructblip}, based on their temporal consistency and differences relative to the static frame. This approach enables effective token compression while integrating multiple modalities within the video context. (3) To enhance the effectiveness of the model on extremely long video, we introduce the \textbf{L}ong \textbf{V}ideo \textbf{C}hain-of-\textbf{T}hought (LVCoT) strategy, which guides the model to process long videos step by step before integrating the whole video to generate the final output.

We train models of various sizes using a multi-stage strategy, progressively optimizing them for vision-language alignment, video instruction tuning, and audio-video instruction tuning.
We evaluate our models on a range of video benchmarks, including MVBench~\cite{li2024mvbench}, PerceptionTest~\cite{patraucean2023perception}, EgoSchema~\cite{mangalam2023egoschema}, MLVU~\cite{zhou2024mlvu}, and Video-MME~\cite{fu2024video-mme}. Furthermore, we assess their performance on audio-video question-answering benchmarks, such as Music-QA~\cite{music-avqa} and AVSD~\cite{alamri2019avsd}. Resuls show that our models achieve strong performance on both video and multimodal understanding tasks. Our main contributions include: 

\begin{itemize}[leftmargin=18 pt, itemsep= 1 pt,topsep = -1 pt]
\item We propose a framework for multimodal video modeling, which represents videos using both static visual features and dynamic multimodal context, effectively integrating visual and audio information within a unified video context. 
\item We introduce the \textbf{L}ong \textbf{V}ideo \textbf{C}hain-of-\textbf{T}hought (LVCoT), a training-free strategy that enables MLLMs to process and reason over long videos step by step, enhancing the performance of existing models.
\item We conduct extensive experiments with MLLMs of various sizes and evaluate them on multiple benchmarks, including general video question answering, long video understanding, and audio-visual video comprehension. Our models achieve strong performance, advancing the field of multimodal long video understanding.
\end{itemize}

\section{Related Work}
\noindent \textbf{Multimodal Large Language Models.}
Previously, multimodal models like CLIP \cite{clip} primarily focused on specific tasks and modalities, including vision and language. Recently, the integration of LLMs and the scaling of data and model size have produced advanced MLLMs. These MLLMs \cite{gpt4v, openai2024gpt4o, team2023gemini, llava, internlm, wang2024internvideo2, llama-adapter} possess powerful understanding, reasoning, and generation capabilities, enabling them to show brilliant performance on a wide range of multimodal tasks.
This proficiency in vision-language understanding can be seamlessly extended to video understanding. For instance, Video-LLaVA~\cite{lin2023video-llava} and LLaMA-VID~\cite{li2024llama-vid} utilize LLMs as decoders to generate answers based on video content and input questions. Several works~\cite{zhang2023videollama, cheng2024videollama2, onellm, geng2024longvale, wang2024internvideo2, internlmxcomposer2_5_OL, openai2024gpt4o, team2023gemini, team2024gemini,  onellm, han2023imagebind-llm} attempt to incorporate additional modalities, such as audio, into LLMs to enhance their perception of the complex world. In addition, some works \cite{zhan2024anygpt, lin2024moma, fei2024vitron, wu24nextgpt} focus on developing a unified MLLM for both understanding and multimodal generation tasks. Vision, language and audio are core components of human perception. To enable models to effectively understand dynamic environments, video comprehension has gained increasing attention. However, it remains a field that requires further exploration.

\begin{figure*}[t]
  \centering
  \includegraphics[width=0.98\linewidth]{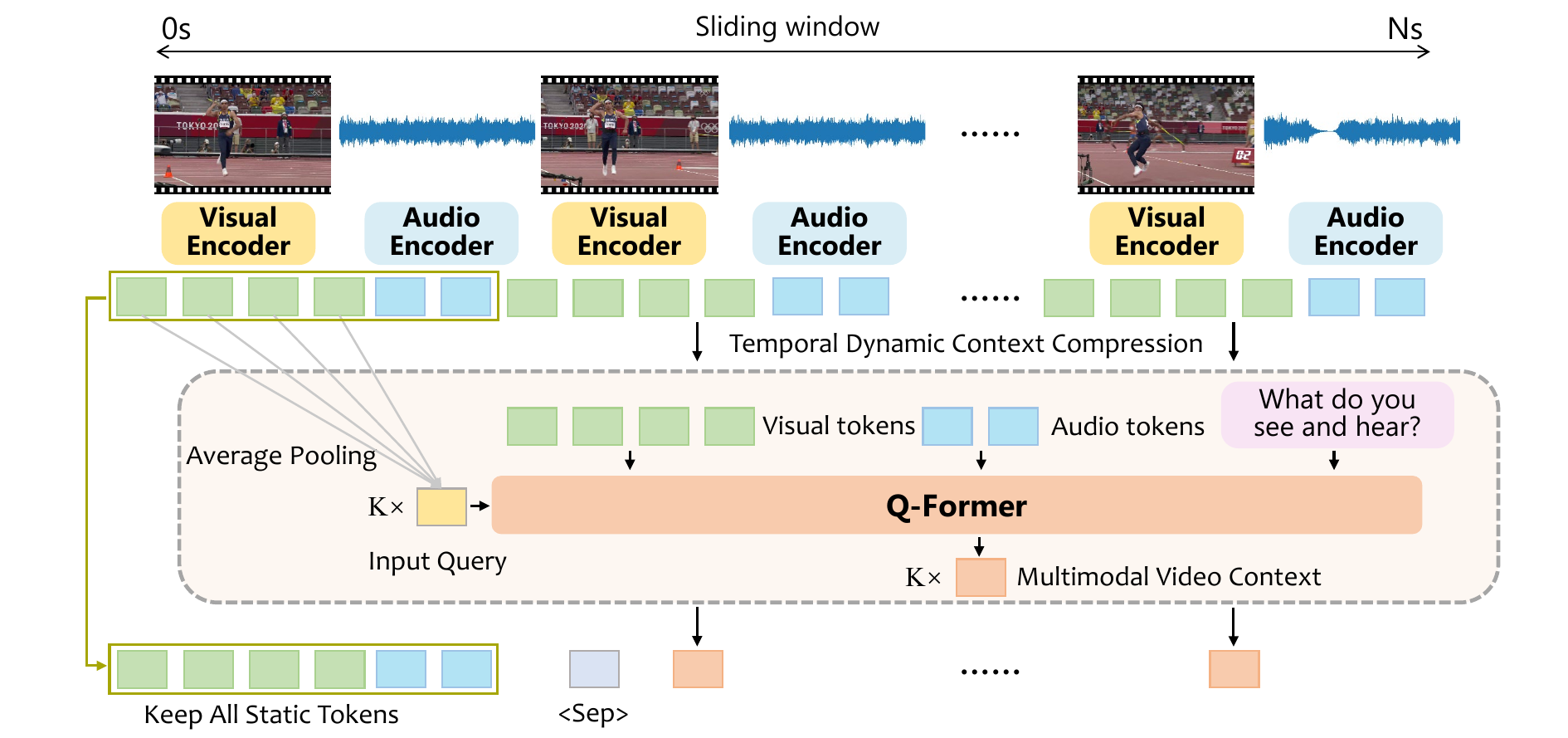}
  \vspace{-0.25cm}
  \caption{\textbf{Architecture of Our Multimodal Video Encoder.} We first extract features for each second of the video, including both visual and corresponding audio tokens. The first frame is selected as the static frame, and a Q-Former is used to perform Temporal Dynamic Context compression based on its relationship with subsequent frames, resulting in $K$ compressed tokens per frame. The final video representation consists of all static frame tokens and multimodal video context.}
  \label{fig:framework}
\end{figure*}

\noindent \textbf{Long Video Understanding.}
The rapid development of MLLMs has enabled researchers to extend their vision-language understanding ability to process videos. Video-LLaVA \cite{lin2023video-llava} and VideoChat \cite{li2023videochat} use video instruction data to train LLMs to generate language responses for input video and questions. However, early video LLMs typically represent a video using multiple images. While this simple representation shows some progress in short video processing, it struggles with longer videos due to significant information loss. 
Recently, some methods \cite{koner2024lookupvit, yang2024visionzip, llavamini} are proposed to compress tokens used for image representation, among which LLaVA-Mini \cite{llavamini} utilizes modality pre-fusion to aggregate visual information into language tokens, which effectively reduces the number of tokens needed for an image. Similar approaches are employed to compress frame tokens in long-video LLMs \cite{li2024llama-vid, wang2025internvideo2-5, yang2024pvc, li2024videochat-flash, longvu, zhang2025tinyllava}. 
Specifically, LLaMA-VID \cite{li2024llama-vid} uses context attention to extract video context relevant to the query. LongVU \cite{longvu} reduces redundant frames based on inter-frame similarity and relevance with text query. VideoChat-Flash \cite{li2024videochat-flash} compresses the visual representation of a video clip by exploiting inter-frame redundancies and semantic correlations.
However, most existing methods focus primarily on vision-based video modeling, which overlook other common modalities, such as audio and speech, that are essential for comprehensive video understanding.

\noindent \textbf{Multimodal Video Modeling.} 
Advancements in MLLMs have made it possible to process videos with multiple modalities. VideoLLaMA2 \cite{cheng2024videollama2} integrates BEATs \cite{chen2023beats} to encode audio for LLM understanding, while PandaGPT \cite{su2023pandagpt} combines ImageBind \cite{girdhar2023imagebind} and Vicuna \cite{chiang2023vicuna} to process six modalities, enabling audio-video-language conversation. NExT-GPT \cite{wu24nextgpt} integrates an LLM with adaptors to perceive multimodal inputs, and uses different diffusion decoders to generate outputs in combinations of text, image, video, and audio. SAVEn-Vid \cite{li2024savenvid} introduces an audio-visual video dataset containing over 58,000 audio-visual instructions and uses it to train an audio-visual MLLM, SAVEnVideo. LongVALE \cite{geng2024longvale} develops an automatic pipeline for unified vision-audio-language-event video annotation and establishes a novel benchmark. However, previous methods typically rely on simple frame sampling for video representation, and straightforward concatenation of different modalities for MLLM understanding. These simplifications limit the effectiveness of long video processing and multimodal integration. In this work, we propose a unified video modeling approach to enhance multimodal long video understanding.

\section{Methodology}
\label{sec:Methodlogy}

In this section, we first introduce the preliminary concepts of LLM-based video understanding in Section~\ref{preliminary}. Next, we present a detailed explanation of our proposed video representation model, Temporal Dynamic Context, in Section~\ref{TDC}.  We then introduce our progressive training strategy to align video-audio information with LLMs in Section~\ref{subsec:training_strategy}. Finally, we propose a training-free chain-of-thought approach to process extremely long videos in Section~\ref{LVCoT}. The architecture of our model is shown in Figure~\ref{fig:framework}.

\subsection{Preliminaries}
\label{preliminary}
The common approach of LLM-based video representation starts by sampling a set of frames from a video and encoding each frame individually using a pretrained visual encoder~\cite{clip}. We denote a video of $T$ seconds at a rate of one frame per second as $\mathbf{X}=\{x_1, x_2, \dots, x_T\}$. 
Previous methods typically sample a fixed number of frames from $\mathbf{X}$, regardless of the video length $T$, \textit{e.g.}, only 8 frames for Video-LLaVA \cite{lin2023video-llava}. 
Each sampled frame is then encoded by the image encoder $\mathcal{E}$ as $F_{x_t}=\mathcal{E}(x_t)$ and projected into a space interpretable by the LLM. 
The resulting tokens from all sampled frames are concatenated with text tokens $F_s$ as input to the LLM. However, a low sampling rate leads to sparse frame selection and significant information loss, while dense sampling results in an excessive number of tokens. This challenge stems from disregarding the temporal relations between frames, limiting the model's ability to process long videos efficiently.

\subsection{Temporal Dynamic Context}
\label{TDC}

Humans process visual input holistically, rather than treating each frame as an independent image. We typically recognize the overall scene first and then focus on dynamic changes.
Inspired by this observation, we propose to model video using both the static features of key frames and the temporal dynamics of the whole scene. Static features allow the model to capture fine-grained visual details, while temporal dynamics encode the evolution of the video over time. In the following section, we introduce a temporal dynamic context encoding method.

\noindent \textbf{Video Scene Segmentation.}
For each video, we maintain the original 1 frame per second (fps) rate to preserve content consistency and prevent temporal information loss. Next, we segment the video into semantically consistent clips based on inter-frame similarities. In contrast, existing methods typically segment videos into fixed-duration clips and encode them separately, neglecting the temporal relations between frames.
We employ the self-supervised vision encoder DINOv2~\cite{oquab2023dinov2}, which is proved to be effective in capturing visual details~\cite{longvu}, to extract high-dimensional embeddings. 
We compute the cosine similarities between consecutive frame pairs and identify the $S{-}1$ points with the lowest frame consistency. Using these points, we segment the input video into $S$ scenes, which enhances temporal coherence in subsequent video encoding.

\noindent \textbf{Static Feature Encoding.} 
For each segmented scene, we represent it using static features along with subsequent dynamic video context. For every second of video, we extract both visual and audio tokens using pretrained vision and audio encoders. Within a sliding window of length $N$, the first frame is selected as the static frame, where all visual and audio tokens are retained in their original form. The remaining frames are then compressed into a temporal dynamic context.

\noindent \textbf{Temporal Dynamic Context.} 
To encode the dynamic evolution of a video, we exploit the relationships between consecutive frames and the static reference frame. 
Previous methods typically compress videos based on visual similarity, overlooking the semantic relationships. This often results in suboptimal compression performance and makes it harder for MLLMs to accurately understand the full video. 
In contrast, we adopt a temporal difference-based strategy by computing the semantic differences between each frame and the static frame, aiming to better preserve meaningful temporal dynamics.
Specifically, we implement it with a Query Transformer~\cite{instructblip} (Q-Former).
We apply average pooling to the static features of the first frame, $F_{x_1}$, to obtain $K$ query tokens $Q\in \mathbb{R}^{K \times D}$ of dimension $D$:
\begin{equation}
Q = \mathrm{AvgPool}(F_{x_1}).
\end{equation}
Note that learnable query tokens are another option, but our experiments (Section~\ref{subsec:ablation}) show that average pooled tokens are more effective.
For each subsequent frame, its visual tokens $F_{x_i}$ and the corresponding audio tokens $F_{a_i}$ are projected to the same dimension and fed to the Q-Former, which performs cross-attention between these tokens and the query tokens:
\begin{align}
F_Q^i = \mathrm{QFormer}(Q,[F_{x_i}\cdot F_{a_i}]), 
\end{align}
where $[\cdot]$ denotes token concatenation.
To make the compression more effective and adaptive to user instructions, we also feed the instruction text $F_s$ into the Q-Former:
\begin{align}
    F_Q^i = \mathrm{QFormer}(Q,[F_{x_i}\cdot F_{a_i}], F_s).
\end{align}
The Q-Former's query output serves as the compressed representation for each frame. These are then concatenated to form the temporal dynamic context $F_{\mathrm{TDC}}$ of the video clip, aggregating both visual and audio information:
\begin{align}
    F_{\mathrm{TDC}} = [F_{x_1}\cdot F_{a_1}\cdot F_Q^2\cdot F_Q^3\cdot \dots\cdot F_Q^N].
\end{align}
This strategy allows the model to selectively allocate attention to specific modalities when answering a given question. Additionally, to differentiate static tokens from dynamic context tokens, we introduce a learnable token, \texttt{<Sep>}, as the separator. Our model tightly integrates visual, audio, and language modalities, providing a promising solution for multimodal video modeling.

\subsection{Multimodal Training Strategy}
\label{subsec:training_strategy}
Multi-stage training has been commonly used and demonstrated effective in previous work \cite{cheng2024videollama2, li2024llama-vid, yang2024pvc}. We train the model in three stages to progressively enhance its understanding of different modalities. In the first stage, we pretrain our model for vision-language alignment using the instruction tuning dataset, LLaVA-OneVision \cite{llavaonevision}. In the second stage, we train the model on a vision-focused video-language dataset without audio. Specifically, we construct the training dataset using LLaVA-Video \cite{zhang2024llava-video}, VideoChat2-IT \cite{li2024mvbench} and MovieChat \cite{moviechat}. In the third stage, we train our model on an audio-visual video understanding dataset to enable the model to comprehend multiple modalities jointly. The data for this stage is collected from Music-AVQA \cite{music-avqa}, AVQA \cite{yang2022avqa}, AVSD \cite{alamri2019avsd}, LongVALE \cite{geng2024longvale} and AVInstruct \cite{ye2024cat}. We also sample a subset of the data used in stage 2 to retain the capabilities learned in previous stages.

\subsection{Long Video Chain of Thought}
\label{LVCoT}
While there have been some advancements in video modeling, processing extremely long videos as a whole is still challenging.
Just like we cannot summarize an entire movie before watching it, MLLMs also need to process long videos progressively. To address this, we propose a method that allows MLLMs to watch and reason through long videos step by step.

Previous approaches \cite{qian2025streaminglong, longvu, ataallah2024goldfish} often rely on key frame selection to answer questions, which disrupts the temporal continuity of the video and makes it more difficult for MLLMs to comprehend the content. Other methods \cite{islam2024videorecap, wei2025longcaptioning} employ a hierarchical strategy, segmenting the video into smaller clips, generating captions for each clip, and then summarizing them to produce a final video description. However, these approaches are typically designed for specific tasks, such as video captioning, and struggle to generalize across different applications. VideoCoT \cite{videocot} introduces active annotation to generate CoT data for training reasoning abilities on videos, but it is limited to short videos. Developing a versatile strategy that can adapt to diverse scenarios remains a significant challenge.
 
To this end, we propose \textbf{L}ong \textbf{V}ideo \textbf{C}hain-of-\textbf{T}hought (LVCoT), a training-free method that can be applied to various MLLMs for extremely long video understanding. We divide the video into multiple time-equivalent segments and query the model to summarize relevant information for answering the given question separately. During this process, the model identifies useful information within each segment and integrates it to generate the final answer. Once the model has processed the entire video, we concatenate all segment outputs along with their corresponding time intervals, representing the model’s thought process. We then query the model to generate the final response based on the global video content. This approach effectively combines segment-level information with the overall context, enabling deeper reasoning.

\input{table/main_result}
\input{table/3b_results}
\input{table/avqa}
\input{table/ablation}

\section{Experiment}
\subsection{Experimental Setup}
We mainly conduct experiments with two backbone LLMs: Qwen2-7B \cite{qwen2} and LLaMA3.2-3B \cite{llama3-2}.
We sample 1 frame per second for each video. Following previous work~\cite{longvu,tong2024cambrian1}, we use DINOv2 \cite{oquab2023dinov2} and SigLIP \cite{zhai2023siglip} as visual encoders, and obtain 144 aggregated tokens per frame. For audio encoding, following the implementation in BEATs \cite{chen2023beats}, we resample the raw audio waveform to 16,000 Hz, and extract audio tokens using the pretrained BEATs encoder, resulting in about 50 tokens per second. We set the maximum number of scene segments to 24, and the number of query tokens to 16 by default. We use the pretrained BERT~\cite{devlin2019bert} to initialize the Q-Former.
More implementation details can be found in the Appendix~\ref{appendix:expetiment}.

\subsection{General Video Understanding}
First, we evaluate models on vision-focused video understanding benchmarks, including MVBench \cite{li2024mvbench}, PerceptionTest \cite{patraucean2023perception}, EgoSchema \cite{mangalam2023egoschema}, MLVU \cite{zhou2024mlvu} and Video-MME \cite{fu2024video-mme}. The results are present in Table \ref{tab:main}. From the results, in short video understanding benchmarks such as MVBench and PerceptionTest, most MLLMs perform well and achieve high accuracy. Longer videos pose greater challenges, leading to a performance drop in almost all models.
Compared to existing audio-visual MLLMs, our model is the first to model dense frames and audios in a unified framework and consistently achieves the best results in video understanding. Notably, models that rely on sparsely sampled frames experience a significant decline in performance on long video benchmarks such as MLVU and VideoMME. In these cases, our model outperforms VideoLLaMA2 by 15.6\% and 9.9\%, respectively. Compared to vision-focused MLLMs, our model also shows competitive performance while being additionally capable of understanding audio within video inputs.

\noindent\textbf{Smaller Model.} In addition, we train a smaller model based on Llama3.2-3B~\cite{llama3-2}. For the sake of data and computational efficiency, we sample a subset of the original data for training, with details provided in  Appendix \ref{appendix:expetiment}. The results are presented in Table \ref{tab:3bresults}. At the parameter scale of 3B-4B, our model achieves the best performance in both short and long video understanding. Notably, with a similar amount of training data, our TDC model outperforms LongVU on MLVU by 7.4\%, further demonstrating its effectiveness.

\begin{figure*}[t]
  \centering
  \includegraphics[width=0.92\linewidth]{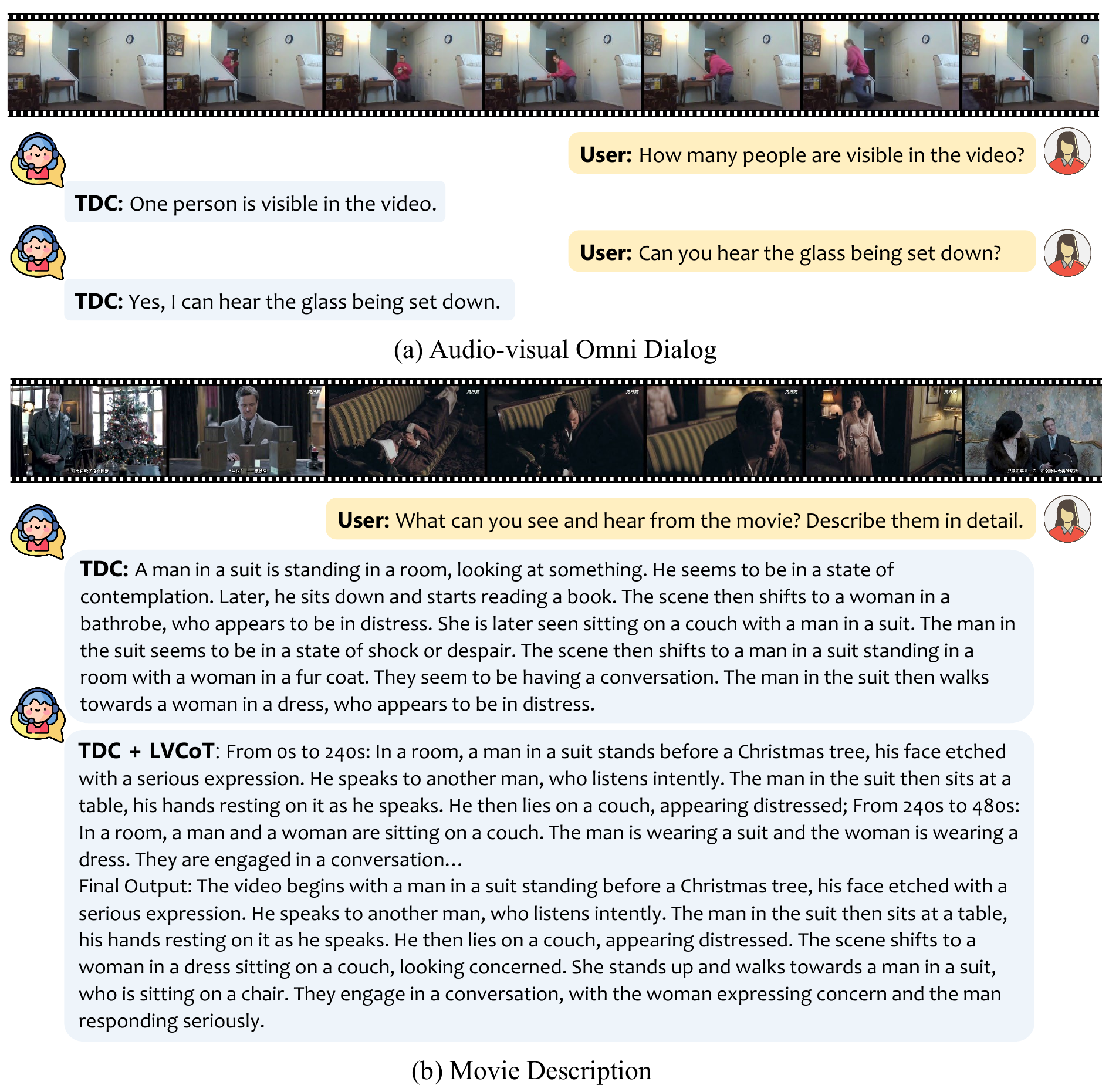}
  \vspace{-0.05cm}
  \caption{\textbf{Qualitative Demonstrations of Our 7B Model.} (a) Our model can uniformly comprehend both audio and visual information, demonstrating strong performance in audio-visual dialogue tasks. (b) In movie description tasks, it can generate detailed descriptions of both the plot and visual elements. For extremely long videos, our LVCoT processes them segment by segment. The generated segment information, along with the timeline, serves as part of the reasoning process, enriching the final output with more details.}
  \label{fig:demo}
\end{figure*}

\subsection{Audio-Visual Omni Video Understanding}
We evaluate models on audio-visual joint video understanding benchmarks, including Music-AVQA \cite{music-avqa}, audio-visual scene-aware dialog (AVSD) \cite{alamri2019avsd}. Music-AVQA contains 9129 samples for evaluating models with visual and audio understanding of musical performance. And AVSD includes 18630 samples of open-ended questions about visual and audio scenes in daily dialogue scenarios. The results are provided in Table \ref{tab:avqa}. Our model achieves the best result on AVSD and shows compatible performance with VideoLLaMA2 on Music-AVQA.

\subsection{Ablation Study}
\label{subsec:ablation}

\noindent\textbf{Effects of Segmentation.}
To evaluate the impact of consistency-based segmentation on video understanding, we vary the maximum number of segments, train the model, and assess its performance. The results are shown in \textbf{(a)} of Table \ref{tab:ablation}. When the maximum is set to one, it means the entire video is processed as a whole, the performance drops remarkably. This is because it incorrectly establishes relationships between non-contiguous video frames, making it difficult for the context tokens to capture the complete video information. This effect is particularly evident in short videos with rapid scene changes. On the other hand, increasing the number of segments to 48 does not result in additional improvements, indicating that our choice of 24 is sufficient to divide the video into appropriate scenes.

\noindent\textbf{Avg Pooling \textit{vs.} Learned Queries.} Learnable queries are commonly used in querying transformers to extract information from different modalities. We compare our model with a variant trained using learnable query tokens, and the results are shown in \textbf{(b)} of Table \ref{tab:ablation}. While learnable queries achieve comparable performance in context compression, they introduce additional computational overhead. In contrast, tokens obtained through average pooling effectively represent the static reference frame, and help extract dynamic changes in subsequent frames. This approach also has the advantage of adaptively adjusting the number of context tokens for dynamic compression.

\noindent\textbf{Number of Context Tokens.}
We conduct experiments with varying numbers of context tokens. The results, presented in \textbf{(c)} of Table \ref{tab:ablation}, indicate that increasing context tokens does not necessarily improve performance. Although more context tokens can capture additional video information, they also increase the number of tokens per frame, thus restrict the number of frames processed and increasing the computational overhead for MLLM, This highlights a trade-off between retaining more information within each frame and encoding a greater number of frames.

\noindent\textbf{Text Instruction in Context Compression.}
We evaluate the contribution of text instructions in video context compression, the results are shown in \textbf{(d)} of Table \ref{tab:ablation}. From the results, we can see that the text instructions help to improve models performance on various dataset. This is because text instructions offer valuable guidance to the compressor in identifying essential information to answer the question, thereby enhancing the efficiency of context compression.

\noindent\textbf{Effects of LVCoT.}
When processing the entire video as a whole, understanding and summarizing useful information in video can be challenging. As shown in Table~\ref{tab:ablation} (e),
applying LVCoT to both 3B and 7B models improves performance on different video benchmarks. Notably, the improvements become more significant as video length increases, demonstrating LVCoT's effectiveness in long video understanding.

\subsection{Qualitative Demonstrations}
In Figure~\ref{fig:demo}, we present several examples demonstrating our model’s general video understanding capabilities. Specifically, Figure \ref{fig:demo} (a) shows how our model uniformly comprehends both audio and visual information, which enhances its ability as a personal assistant. Figure \ref{fig:demo} (b) showcases its performance in movie understanding, where it generates detailed descriptions of both the plot and visual elements. For extremely long videos, such as movies, our LVCoT processes them segment by segment, further improving the quality of the descriptions.

\section{Conclusion}
In this paper, we introduce a novel multimodal long-video modeling framework named \textbf{T}emporal \textbf{D}ynamic \textbf{C}ontext (TDC). This framework represents a video using both static visual features and dynamic video context within each scene, which provide visual details and dynamic motions of the video, respectively. Our model integrates multiple modalities into a unified video context, enhancing multimodal joint long-video understanding. For extremely long video, we introduce the \textbf{L}ong \textbf{V}ideo \textbf{C}hain-of-\textbf{T}hought (LVCoT) strategy, which guides the model to process long videos step by step before integrating the full video to generate the final output. This approach improves model performance and allows models with limited context windows to effectively handle longer videos. Extensive experiment demonstrate that our model achieves strong performance across general video understanding tasks and audio-visual omni video understanding benchmarks.





%% file: table/main_result.tex
\begin{table*}[t]
\centering
\begin{adjustbox}{width=\linewidth,center}
\renewcommand{\arraystretch}{1.1}
\setlength{\tabcolsep}{1.5mm}
\begin{tabular}{lrccccccc}
\toprule  {\textbf{Model}} & \multicolumn{1}{c}{{\centering \textbf{Size}}} & {\textbf{\#Frames}}& {\textbf{\#Tokens}}  & {\textbf{MVBench}} & {\textbf{PerceptionTest } } & {\centering \textbf{EgoSchema} }  & {\textbf{MLVU}} & {\centering \textbf{VideoMME}}   \\
Average duration (sec) & & & \textbf{per Frame}& 16 & 23 & 180 & 651 &  1010\\
\midrule
\textit{Commercial Models} \\
GPT4-V~\citep{gpt4v} & - & 1fps&- & 43.7 & - & -   & 49.2 & 59.9 \\
GPT4-o~\citep{openai2024gpt4o} & - &1fps& - & 64.6 & - & 72.2 & 64.6 & 71.9 \\
Gemini-1.5-Pro~\cite{team2024gemini} & - &1fps& - & 60.5 & - & 71.2  & - & 75.0 \\
\midrule
\textit{Vision-focused MLLMs} \\
InternVL2~\cite{internvl2} & 8B & 12 &256 & 66.4 & - &  -  & - & 54.0 \\
LLaVA-NeXT-Video~\cite{llavanextvideo} & 7B & 32 & 144 & 53.1 & 48.8 & -  & - & 46.5  \\
LLaVA-OneVision~\cite{llavaonevision} & 7B &32 &196 &56.7 & 57.1  & 60.1  & 64.7 & 58.2 \\
\color{gray} LLaVA-OneVision~\cite{llavaonevision} & \color{gray} 72B & \color{gray}32& \color{gray}196 &\color{gray}59.4 & \color{gray}66.9  & -  & \color{gray}68.0 & \color{gray}66.2 \\
mPLUG-Owl3~\cite{ye2024mplug} & 7B & 8 & - & 54.5 & -  & - & - & 53.5\\
Qwen2-VL~\cite{Qwen2-VL} & 7B & 2fps & - & 67.0 & 62.3  & 66.7 & -  & 63.3\\
VideoChat2-HD~\cite{li2024mvbench} & 7B & 16 & 72 & 62.3 & - & -  & 47.9 & 45.3 \\
InternVideo2-HD~\citep{wang2024internvideo2} & 7B & 16 & 72 & 67.2 & 63.4  & 60.0 & -  & 49.4 \\
VideoChat-TPO \cite{yan2024task} & 7B & 16 & 64 & 66.8 & - & - & 54.7 & - \\
InternVL2.5 \cite{wang2025internvideo2-5} & 7B & 12 & 256 & 72.0 & 68.2 & 51.5   & 68.9 & 64.2 \\
LLaMA-VID~\citep{li2024llama-vid} & 7B & 1fps & 2 & 41.9 & 44.6 & - & 33.2 & 25.9\\
LongVILA~\citep{longvila} & 7B & 2048 & 196 & - & - & 67.7  & - & 57.5 \\
LongVA~\citep{longva} & 7B & 128 & 144 & - & - & - & 56.3 & 52.6 \\
LongLLaVA~\citep{longllava} & 9B & 128 & 144 & 49.1 & - & - & -& 43.7 \\
LLaVA-Video~\citep{longvu} & 7B & - & 169& 58.6 & 67.9 & 57.3  & 70.8 & 63.3 \\
LongVU~\citep{longvu} & 7B & 1fps & 144/64 & 66.9 & - & 67.6 & 65.4 & - \\
PVC\textsubscript{InternVL2}~\citep{yang2024pvc} & 8B &96 & 64 & 73.8 & 68.4 & 59.6 & 72.4 & 64.1 \\
MAmmoTH-VL~\citep{guo2024mammothvl} & 8B & 5 & 729 & 59.1 & 59.3 & 58.5 & 64.7 & 58.8 \\
\midrule
\textit{Audio-visual MLLMs} \\
PandaGPT~\cite{su2023pandagpt} & 7B & 10 & 196 & - & - & - & -  & 43.5   \\
NExT-GPT~\cite{wu24nextgpt} & 7B & 24 & 196 & - & - & - & -  & 42.6  \\
VideoLLaMA2~\citep{cheng2024videollama2} & 7B & 16 &72 & 54.6 & 51.4 & 51.7  & 48.5 & 47.9 \\
\color{gray}VideoLLaMA2~\citep{cheng2024videollama2} & \color{gray} 72B &\color{gray} 16&\color{gray} 72 & \color{gray}62.0 & \color{gray}57.5 & \color{gray}63.9  &\color{gray}- & \color{gray}61.4 \\
VideoLLaMA2.1\cite{cheng2024videollama2} & 7B & 16& 72 & 57.3 & 54.9 & 53.1 &  - & 54.9\\
\rowcolor{lightgray!20} \textbf{TDC (Ours)} & 7B & 1fps & 16 & \textbf{68.3} & \textbf{67.5} & \textbf{65.7} & \textbf{64.1}  & \textbf{57.8} 
\\
\bottomrule
\end{tabular}
\end{adjustbox}
\caption{\textbf{Results on Video Question Answering Benchmarks,} including short video and long video understanding. We compare our model with Vision-focused MLLMs and Audio-visual Omni MLLMs.
We present the performance of our model with the proposed LVCoT. The best results among Audio-visual MLLMs are bold. Results on VideoMME are evaluated without subtitles.
}
\label{tab:main}
\end{table*}

%% file: table/3b_results.tex
\begin{table*}[t]
    \centering
\begin{adjustbox}{width=1\linewidth,center}
\renewcommand{\arraystretch}{1.1}
\setlength{\tabcolsep}{1.5mm}
\begin{tabular}{lccccccccc}
\toprule  {\textbf{Model}} & \multicolumn{1}{c}{{\centering \textbf{LLM}}}  & \multicolumn{1}{c}{{\centering \textbf{Size}}} & {\textbf{\#Frames}} & {\textbf{\#Tokens}}  & {\textbf{MVBench}}  & {\centering \textbf{EgoSchema} }  & {\textbf{MLVU}} & {\centering \textbf{VideoMME}}  \\
Average duration (sec) & &  & &\textbf{per Frame} & 16 & 180 & 473 &  1010 \\
\midrule
\textit{Vision-focused MLLMs} \\
InternVL2~\cite{internvl2} & InternLM2 \cite{internlm}&1.8B & 16 & 256 & 60.2 & - & 47.3 & - \\
VideoChat2~\cite{li2024mvbench} & Phi-3-mini \cite{abdin2024phi}& 4B & 16 & 96 & 55.1 & 56.7 & - & - \\
Phi-3.5-vision-instruct~\cite{abdin2024phi} &Phi-3-mini \cite{abdin2024phi}& 4B & 16 & 256 & - & 50.8 & - & - \\
TinyLLaVA-Video~\cite{zhang2025tinyllava} &Qwen2.5 \cite{qwen2.5}& 3B &16 & - & 42.5 & - & 48.1 & -  \\
LongVU~\cite{longvu} & Llama3.2 \cite{llama3-2}&3B & 1fps&144/64 & 60.9 & 59.1 & 51.5 & 55.9 \\
\midrule
\textit{Audio-visual MLLMs} \\
\rowcolor{lightgray!20} \textbf{TDC (Ours)} & Llama3.2 \cite{llama3-2}&3B & \textbf{1fps} &\textbf{16} & \textbf{62.7} & \textbf{61.0} & \textbf{58.9} &\textbf{59.5} \\
\bottomrule
\end{tabular}
\end{adjustbox}
\caption{\textbf{Results of Smaller Sized Models. } 
We present the performance of our model with the proposed LVCoT. Results on VideoMME are evaluated with subtitles. The best results are bold.
}
\label{tab:3bresults}
\end{table*}

%% file: table/avqa.tex
\begin{table}[t]
\centering
\begin{adjustbox}{width=1.0\linewidth,center}
\renewcommand{\arraystretch}{1.35}
\setlength{\tabcolsep}{1.35mm}
\begin{tabular}{lccccc}
\toprule
\textbf{Model}                                                                & \textbf{Size} & \textbf{\#Frames} & \textbf{\#Tokens} & \textbf{AVSD} & \textbf{Music-AVQA} \\ \midrule
PandaGPT~\cite{su2023pandagpt}                          & 13B           & 10                & 196               & 26.1          & 33.7                \\
NExT-GPT~\cite{wu24nextgpt}                             & 7B            & 24                & 196               & -             & 79.8                \\
VideoLLaMA2~\citep{cheng2024videollama2}                & 7B            & 16                & 72                & 57.2          & 79.2                \\
VideoLLaMA2.1~\cite{cheng2024videollama2}                     & 7B            & 16                & 72                & 57.2          & 80.9                \\
LongVALE~\cite{geng2024longvale}                              & 7B            & 100               & 256               & 54.8          & 49.4                \\
\rowcolor{lightgray!20}\textbf{TDC (Ours)} & 7B            & \textbf{1fps}              & \textbf{16}                & \textbf{57.6}          & 78.7                \\ \bottomrule
\end{tabular}
\end{adjustbox}
\caption{\textbf{Results on Audio-Visual Omni Video Understanding}, including AVSD~\cite{alamri2019avsd} and Music-AVQA~\cite{music-avqa}.}
\label{tab:avqa}
\vspace{-3mm}
\end{table}

%% file: table/ablation.tex
\begin{table}[t]
\centering
\begin{adjustbox}{width=1\linewidth,center}
\renewcommand{\arraystretch}{1.2}
\setlength{\tabcolsep}{1.5mm}
\begin{tabular}{lllll}
\toprule
Dataset  & MVBench & MLVU & \multicolumn{2}{c}{VideoMME} \\
& & & Overall & Long \\
\midrule
\multicolumn{5}{l}{\bf (a) Maximum Number of Segments}         \\
\midrule
1 (No Segment)           &  53.5 {\footnotesize(\textcolor{Red}{-9.2})}  & 56.6{\footnotesize(\textcolor{Red}{-1.7})}  &  58.7 {\footnotesize(\textcolor{Red}{-0.9})} & 53.2 {\footnotesize(\textcolor{Green}{+0.5})}\\
\rowcolor{lightgray!20}
24           &  62.7 & 58.3 &   59.6  &  52.7 \\
48           & 62.2 {\footnotesize(\textcolor{Red}{-0.5})} & 58.3  &  58.5{\footnotesize(\textcolor{Red}{-1.1})} & 51.0 {\footnotesize(\textcolor{Red}{-1.7})}  \\
\midrule
\multicolumn{5}{l}{\bf (b) Query Type}             \\
\midrule
Learned Query          & 61.7{\footnotesize(\textcolor{Red}{-1.0})} & 58.2{\footnotesize(\textcolor{Red}{-0.1})}      & 59.5{\footnotesize(\textcolor{Red}{-0.1})}    &   52.1{\footnotesize(\textcolor{Red}{-0.6})}    \\
\rowcolor{lightgray!20}
AvgPooling      &  62.7  & 58.3 &   59.6  &  52.7 \\
\midrule
\multicolumn{5}{l}{\bf (c) \#Context Tokens per Frame}\\
\midrule
32   &  61.7{\footnotesize(\textcolor{Red}{-1.0})} & 56.1{\footnotesize(\textcolor{Red}{-2.2})} & 58.4 {\footnotesize(\textcolor{Red}{-1.2})}  &   52.1 {\footnotesize(\textcolor{Red}{-0.6})}   \\
\rowcolor{lightgray!20}
16       &  62.7  & 58.3  &   59.6   &  52.7  \\
\midrule
\multicolumn{5}{l}{\bf (d) Text Instruction}      \\
\midrule
Without Text.  & 62.3{\footnotesize(\textcolor{Red}{-0.4})} & 58.1{\footnotesize(\textcolor{Red}{-0.2})} &   58.0{\footnotesize(\textcolor{Red}{-1.6})}  & 51.5{\footnotesize(\textcolor{Red}{-1.2})}  \\
\rowcolor{lightgray!20}
Text Input.     &  62.7  & 58.3  &   59.6   &  52.7  \\
\midrule
\multicolumn{5}{l}{\bf (e) Effect of LVCoT} \\
\midrule
\rowcolor{lightgray!20} 3B &  62.7  & 58.3  &  59.6    &  52.7   \\
3B \textit{w/} LVCoT & 62.7 & 58.9{\footnotesize(\textcolor{Green}{+0.6})} & 59.5{\footnotesize(\textcolor{Red}{-0.1})} & 52.7 \\
\rowcolor{lightgray!20} 7B & 68.3 & 63.9 & 65.9 & 61.3 \\
7B \textit{w/} LVCoT & 68.3 & 64.1{\footnotesize(\textcolor{Green}{+0.2})} & 66.2{\footnotesize(\textcolor{Green}{+0.3})} & 61.8 {\footnotesize(\textcolor{Green}{+0.5})} \\
\bottomrule
\end{tabular}
\end{adjustbox}
\vspace{-1mm}
\caption{\textbf{Results of Ablation Studies.} We conduct ablation studies on: (a) the maximum number of scene segments in video encoding, (b) the type of query used for temporal context compression, (c) the number of context tokens for each frame, (d) the effect of text information in context compression,
(e) the effect of LvCoT with 3B and 7B models. The row with a \colorbox{lightgray!20}{gray background} indicates our default setting.}
\vspace{-3mm}
\label{tab:ablation}
\end{table}

%% file: sec/X_suppl.tex
\clearpage

\appendix

\section{Appendix Overview}
\begin{itemize}[leftmargin=18 pt, itemsep= 3 pt,topsep = 1pt]
\item Section \ref{appendix:eval}: Additional evaluations of our models.
\item Section \ref{appendix:expetiment}: More experimental details.
\item Section \ref{appendix:limitation}: Analysis on limitations of our work.
\end{itemize}

\section{Additional evaluations}
\label{appendix:eval}
\input{table/appendix_eval}
In Table \ref{tab:appendix_results}, we provide a more detailed comparison on the VideoMME \cite{fu2024video-mme} dataset. In this evaluation, subtitles for each video are provided to the model. The results show that our model consistently achieves the best performance across both short and long video settings, which demonstrates its adaptability to a wide range of video scenarios.

\section{Experimental details}
\label{appendix:expetiment}

\subsection{Training data}
Our training process contains three stage. In the first stage, we pretrain our model on vision-language alignment using the single image instruction tuning dataset, LLaVA-OneVision \cite{llavaonevision}. In the second stage, we train our model on a vision-focused video-language dataset without audio. Specifically, we construct the training dataset using LLaVA-Video \cite{zhang2024llava-video}, VideoChat2-IT \cite{li2024mvbench} and MovieChat \cite{moviechat}. In the third stage, we train our model on an audio-visual video understanding dataset to enable the model to comprehend multiple modalities jointly. Our training data is collected from Music-AVQA \cite{music-avqa}, AVQA \cite{yang2022avqa}, AVSD \cite{alamri2019avsd}, LongVALE \cite{geng2024longvale} and AVInstruct \cite{ye2024cat}. We also sample a subset from the data used in stage 2 to retain the capabilities learned in previous stages. The detailed data sources are listed in Table \ref{tab:datasets}.

\subsection{Implementation details}
We mainly conduct experiments with two backbone LLMs: Qwen2-7B \cite{qwen2} and LLaMA3.2-3B \cite{llama3-2}.
We sample 1 frame per second for each video. Following previous work~\cite{longvu,tong2024cambrian1}, we use DINOv2 \cite{oquab2023dinov2} and SigLIP \cite{zhai2023siglip} as visual encoders, and obtain 144 aggregated tokens per frame. For audio encoding, following the implementation in BEATs \cite{chen2023beats}, we resample the raw audio waveform to 16,000 Hz, and extract audio tokens using the pretrained BEATs encoder, resulting in about 50 tokens per second. We set the maximum number of scene segments to 24, and the number of query tokens to 16 by default. We use the pretrained BERT~\cite{devlin2019bert} to initialize the Q-Former. 

The models are trained for one epoch in each stage. During training, the visual and audio encoders are kept frozen, while the temporal compressor and the MLLMs are trained. In the first two stages, we train the full model parameters. In the third stage, we apply Low-Rank Adaptation (LoRA) \cite{lora} to reduce GPU memory consumption. The detailed hyperparameter settings used during model training are presented in Table \ref{tab:hyper}.

\setlength{\tabcolsep}{2pt}
\setlength{\doublerulesep}{2\arrayrulewidth}
\renewcommand{\arraystretch}{1.15}

\setlength{\tabcolsep}{2pt}
\setlength{\doublerulesep}{2\arrayrulewidth}
\renewcommand{\arraystretch}{1.15}

\begin{table}[h]
    \renewcommand{\thetable}{5}
    \centering
    \small
    \resizebox{0.85\linewidth}{!}{
    \begin{tabular}{lccc}
        \toprule
        Training Stage & Stage 1 & Stage 2 & Stage 3 \\
        \midrule
        Max Sequence Length & \multicolumn{3}{c}{8192} \\
        Number of Video Frames & \multicolumn{3}{c}{1 fps} \\
        Number of Segmented Scenes & \multicolumn{3}{c}{24} \\
        Visual Tokens per Frame & \multicolumn{3}{c}{144} \\
        Audio Tokens per Frame & \multicolumn{3}{c}{50} \\
        Context Tokens per Frame & \multicolumn{3}{c}{16} \\
        Optimizer & \multicolumn{3}{c}{AdamW~\cite{loshchilov2017decoupled}} \\
        Learning Rate & 1e-5 & 1e-5 & 2e-5 \\
        Learning Rate Schedule & \multicolumn{3}{c}{Cosine Decay} \\
        Warmup Ratio & \multicolumn{3}{c}{0.03} \\
        Training Mode & Full & Full & LoRA \\
        \bottomrule
    \end{tabular}
    }
    \caption{\textbf{Hyperparameters Used in Model Training.}}
    \label{tab:hyper}
    \vspace{-1em}
\end{table}

\subsection{Evaluation setup}
Following the approach in \cite{cheng2024videollama2}, we adopt an LLM assisted evaluation for AVSD. We also provide an example as one shot. For LVCoT, we set the number of segments to 3 by default.

\begin{table*}[t!]
\centering
\renewcommand{\arraystretch}{1.4}
\setlength{\tabcolsep}{1.5mm}
\footnotesize
\begin{tabular}{y{130}|x{75}y{200}}
\toprule
\textbf{Training stage} & \textbf{\# Samples} & \textbf{Data Sources} \\
\hline
Stage1: Vision-Language Pre-training    & 3.2M & LLaVA-OneVision \cite{llavaonevision}  \\
\midrule
Stage2: Video Instruction Tuning &\begin{tabular}[c]{@{}l@{}}Qwen2-7B: 2M\\ LLama3.2-3B: 540K 
\end{tabular}& \begin{tabular}[c]{@{}l@{}}LLaVA-Video \cite{zhang2024llava-video}, TextVR \cite{wu2025large}, YouCook2 \cite{zhou2018towards}, EgoQA~\cite{fan2019egovqa},\\Kinetics-710~\cite{kay2017kinetics}, NExTQA~\cite{xiao2021next}, CLEVRER~\cite{yi2019clevrer}, TGIF~\cite{li2016tgif}, \\WebVidQA~\cite{yang2021just}, DiDeMo~\cite{anne2017didemo}, ShareGPT4Video \cite{chen2024sharegpt4video},\\ MovieChat \cite{moviechat}
\end{tabular} \\
\midrule
Stage3: Audio-Video Instruction Tuning & \begin{tabular}[c]{@{}l@{}}Qwen2-7B: 300K\\ LLama3.2-3B: 120K
\end{tabular} & \begin{tabular}[c]{@{}l@{}}AVQA \cite{yang2022avqa}, Music-AVQA \cite{music-avqa}, AVSD \cite{alamri2019avsd}, LongVALE \cite{geng2024longvale}, \\
AVinstruct \cite{ye2024cat}, subset from Stage 2
\end{tabular} \\
\bottomrule
\end{tabular}
\vspace{-1mm}
\caption{Datasets used in multi-stage multimodal training.}
\label{tab:datasets}
\end{table*}

\section{Limitations}
\label{appendix:limitation}
The effectiveness of LVCoT depends on the reasoning ability of the MLLM, since the model has not been trained on this task, the improvement is relatively small. In the future, we will explore training the model to better utilize this strategy. Additionally, processing videos multiple times incurs additional computational costs. It would be promising to explore new methods for establishing more efficient memory mechanisms in MLLMs to enhance long video understanding.

%% file: table/appendix_eval.tex
\begin{table}[h]
\centering
\small
\begin{adjustbox}{width=\linewidth,center}
\renewcommand{\arraystretch}{1.1}
\setlength{\tabcolsep}{1.5mm}
\begin{tabular}{lcccccc}
    \toprule
        \textbf{Model} & \textbf{Size} & \textbf{Frames} & \textbf{S} & \textbf{M} & \textbf{L} & \textbf{Overall} \\
        \midrule
        Video-LLaVA \cite{lin2023video-llava} & 7B & 8 & 46.1 & 40.7 & 38.1 & 41.6 \\
        ShareGPT4Video~\citep{chen2024sharegpt4video} & 8B & 16 &  53.6 & 39.3 & 37.9  & 43.6 \\
        Chat-Univi-v1.5~\citep{jin2023chatunivi} & 7B & 64 & 51.2 &  44.6  & 41.8  & 45.9 \\
        VideoLLaMA2~\citep{cheng2024videollama2} & 7B & 16 & 59.4 & 47.6  & 43.8 & 50.3 \\
        VideoChat2~\citep{li2024mvbench} & 7B & 16  & 52.8 & 39.4  & 39.2 & 43.8 \\ 
        LongVA~\citep{longva} & 7B & 128 & 61.6 & 50.4  & 47.6 & 54.3 \\
        LLaVA-OneVision~\cite{llavaonevision} & 7B & 32 & 69.1 & 53.3 & 46.7 & 58.2 \\
        LongVU \cite{longvu}& 7B & 1fps & 64.7	& 58.2 & 59.5 & 60.9 \\
        \rowcolor{lightgray!20} \textbf{TDC (Ours)} & 7B & 1fps       &  \textbf{70.0}  & \textbf{66.2}  &   \textbf{61.3}   &  \textbf{65.9}  \\
        \bottomrule
    \end{tabular}
\end{adjustbox}
\caption{\textbf{Detailed Results on VideoMME.} The best results are bold. Subtitles of videos are provided in this evaluation. S: Short. M: Medium. L: Long.
}
\label{tab:appendix_results}
\end{table}